\documentclass[10pt]{article}

\usepackage{amsmath}
\usepackage{amssymb}

\usepackage{graphicx}

\usepackage{cite}

\usepackage{color} 

\usepackage[labelfont=bf,labelsep=period,justification=raggedright]{caption}

\makeatletter
\renewcommand{\@biblabel}[1]{\quad#1.}
\makeatother

\date{}

\begin{document}

\begin{flushleft}
{\Large
\textbf{High quality topic extraction from business news explains abnormal financial market volatility}
}
\\
Ryohei Hisano$^{1,2}$, 
Didier Sornette$^{1,3}$, 
Takayuki Mizuno$^{2,4,5}$, 
Takaaki Ohnishi$^{2,5}$, 
Tsutomu Watanabe$^{2,5}$
\\
\bf{1} ETH Zurich, Department of Management, Technology and Economics, Zurich, Switzerland
\\
\bf{2} The Canon Institute for Global Studies, Tokyo, Japan
\\
\bf{3} Swiss Finance Institute, Switzerland
\\
\bf{4} University of Tsukuba, Doctoral Program in Computer Science, Graduate School of Systems and Information
Engineering, Ibaraki, Japan
\\
\bf{5} The University of Tokyo, Graduate School of Economics, Tokyo, Japan 
\\
$\ast$ E-mail: em072010@yahoo.co.jp
\end{flushleft}

\section*{Abstract}
Understanding the mutual relationships between information flows and social activity in society today is one of the cornerstones of the social sciences. In financial economics, the key issue in this regard is understanding and quantifying how news of all possible types (geopolitical, environmental, social, financial, economic, etc.) affect trading and the pricing of firms in organized stock markets. In this article, we seek to address this issue by performing an analysis of more than 24 million news records provided by Thompson Reuters and of their relationship with trading activity for 206 major stocks in the S\&P US stock index. We show that the whole landscape of news that affect stock price movements can be automatically summarized via simple regularized regressions between trading activity and news information pieces decomposed, with the help of simple topic modeling techniques, into their ``thematic'' features. Using these methods, we are able to estimate and quantify the impacts of news on trading. We introduce network-based visualization techniques to represent the whole landscape of news information associated with a basket of stocks. The examination of the words that are representative of the topic distributions confirms that our method is able to extract the significant pieces of information influencing the stock market. Our results show that one of the most puzzling stylized fact in financial economies, namely that at certain times trading volumes appear to be ``abnormally large,'' can be partially explained by the flow of news.  In this sense, our results prove that there is no ``excess trading,'' 
when restricting to times when news are genuinely novel and provide relevant financial information.

\section*{Introduction}
Neoclassical financial economics based on the ``efficient market hypothesis'' (EMH) considers price movements as almost perfect instantaneous reactions to information flows. Thus, according to the EMH, price changes simply reflect exogenous news. Such news - of all possible types (geopolitical, environmental, social, financial, economic, etc.) - lead investors to continuously reassess their expectations of the cash flows that firms' investment projects could generate in the future. These reassessments are translated into readjusted demand/supply functions, which then push prices up or down, depending on the net imbalance between demand and supply, towards a fundamental value. As a consequence, observed prices are considered the best embodiments of the present value of future cash flows. In this view, market movements are purely exogenous without any internal feedback loops. In particular, the most extreme losses occurring during crashes are considered to be solely triggered exogenously. 

The problem with this paradigm is that, in practice, relating actual price movements to particular news has been strikingly elusive. Many attempts to relate price changes to news, be it low frequency or high frequency, have failed to find convincing supportive evidence for the EMH \cite{4,5,6,7,8,9}.  Moreover, it has long been recognized that prices move much too large an extent and trading volume is much too large compared with what would be predicted from the EMH \cite{1,2,3}. This suggests that there is more to price dynamics than just the exogenous flow of information.  Against this background, the concept of ``reflexivity'' has been introduced \cite{Soros}, which embodies the notion that past actions of investors also significantly influence present decisions so as to create feedback loops and significant endogenous dynamics \cite{FilimonovSornette2012}. The unresolved issue until now is then to disentangle exogenous and endogenous factors and understand which news are really important and how they are incorporated in prices. Given the a priori foundational nature of news flows on price formation in financial economics on the one hand and the absence of empirical support for it on the other hand, without such an understanding and the corresponding control that should derive from it, financial markets will remain vulnerable to the excess volatility, wild price swings, bubbles and crashes that have plagued them in recent years as well as over most of their history \cite{ReinhartRogoff}. 
  
The present article represents an attempt to break the above stalemate by (i) using a huge database of business news gathered for institutional investors and (ii) introducing a new methodology to extract relevant news that influence trading activity. This new methodology allows us to remove in large part the endogenous components of price dynamics and to identify a hierarchy of important news. Our approach differs in several important dimensions from the ones employed by previous studies investigating the impact of news on financial markets, such as \cite{10,11,12,13,14,16}. One class of previous studies analyzed the information provided by news only in an aggregated manner without taking into account the specific information content. However, as casual observation indicates, each news record has different meaning to investors and thus different impact on prices, so that just counting the total number of news records for a particular period would not work well. Other previous studies only considered a small restricted set of news, such as earnings reports and the release of new economic data, and thus suffered from the serious limitation of neglecting the possible significant impact of other types of news arriving at the time. One way to circumvent the latter problem could be to use very short time intervals \cite{17} so as to minimize attribution errors. But recent studies, including \cite{Mizuno}, have shown that the impact of news persists over days, weeks and sometimes months, making it difficult, if not impossible, to extract their influence by just using temporal partitioning. 

We address all these problems by performing a simultaneous disaggregated estimation of
the relevant news types with respect to financial trading activity. We mine raw texts of more than 24 million news records provided by Thompson Reuters and examine their impact on trading activity in stocks of the 206 firms listed in the S\&P 500 US stock index for each of which there were more than 5,000 news records over the period from January 2003 to June 2011. To determine what pieces of information are the most relevant to explain trading activity of each stock, we use a combination of regularized regressions and topic modeling techniques. This allows us to compare quantitatively the relative importance of the different news. We show that nearly 30-40\% of the top 5\% most important events in terms of trading volume can be almost perfectly explained by our decoded news flow.

\section*{Methods}

The existence of a good correspondence between the time evolution of trading activity 
(measured by the daily trading volume) and the time evolution of news volume is well-known
\cite{10,11,12}. This correspondence is illustrated in Fig. 1, which shows the time evolution of the trading volume (the number of shares traded per day) of the Toyota stock
and the evolution of the volume of news, measured as the number of words per day in text records that include the company name Toyota.  Using  just the number of news records 
(instead of the total number of words in these records) yields essentially the same results.

Starting from this rough aggregate correspondence, our much more ambitious goal is to disaggregate (a) the flow of news into relevant topics and their associated words and (b) the trading volume of individual stocks, in order to construct a complete network of interdependences. Fig. 2 provides a flowchart of our methodology, which consists of
(i) decomposing the total flow of news  into their thematic features by applying topic modeling techniques, (ii) estimating their impact on trading activity simultaneously in order to prune out the unimportant topics, and (iii) quantifying how many of the peaks in trading activity can be explained by news shocks.

Once a term (for instance Toyota) is chosen and the associated news records are collected (step (1)), the second step is to decompose news information pieces into their ``thematic'' features, as shown in Fig. 2. This is done by applying a simple topic modeling technique called Latent Dirichlet Allocation (LDA) \cite{18,19}.  Topic models are graphical models \cite{20} which assume that shared global multinomial word distributions (i.e., topic distributions) govern the corpus. Word frequencies within a given document are created from a mixture of these global topic distributions. LDA is the simplest topic model and uses the Dirichlet prior in order to ensure sparsity in the underlying multinomial distribution. This makes learned topics easier to interpret. Since LDA has already yielded  excellent results, we did not find it useful to employ more elaborate topic models. 
We removed common stop words from the original news records and ran LDA by setting 
the number of topics to 100 for all stocks analyzed in this article.  Varying the number of topics according to the number of news records for each stock did not change the result significantly. We used the fast implementation of Smola and Narayanamurthy \cite{21}.

In what follows, we use the news volume $I_k(t)$ of a given topic $k$, which is defined 
as the total number of words tagged  with topic number $k$ on day $t$,
\begin{equation}
I_k(t) = \sum^{}_{d \in I(t)}\sum^{}_{w}N(d,w,k)~,
\label{wbtwr}
\end{equation} 
where $N(d,w,k)$ is the number of times a word $w$ tagged with topic $k$ appeared in document $d$ and 
$I(t)$ is the indicator function of the set of documents on day $t$.  
Fig. 3 presents some examples of the time evolution of the news volume for four
 topics for the term ``Toyota.'' It also lists the top three words of the corresponding topic distributions. A full description is provided in the supporting information.

The fundamental characteristic of LDA (and of topic modeling in general) is that every word that 
appears in the corpus is tagged with a specific topic and is thus assumed to be generated
by the corresponding specific topic distribution. Put differently, even though words in a given document 
can be generated by a mixture of topics, each word is 
assumed to be drawn from exactly one topic.  This procedure makes the interpretation
of the estimated topics easier to comprehend \cite{22}. As highlighted by \cite{23}, this construction, however, has the following negative consequence: because news records, such as ours, have many repeated phrases such as ``double click for more information,'' ``Reuters messaging net,'' or ``top news,'' many topic distributions simply reflect these repeated phrases. One way to deal with this problem is to eliminate these repeated phrases where they 
appear in the original corpus.  However, because it is difficult to construct an algorithm 
that would work well for all the variations found in the huge amount of news records
analyzed here, we chose to prune the topics using topic distributions, employing the following procedure. 
For each topic, we 
focused on the top 6 words of the corresponding topic distribution and eliminated
that topic if these top 6 words were included in the set of words in the  unwanted repeated phrases (Step 2-b in Fig. 1).  
We also removed all topics that appear for less than 80 days (out of the 3103 days 
from January 2003 to June 2011). This excludes topics such as specific symbols and numbers reported in short time intervals.  We also eliminated topics that describe stock market activity, i.e., which
include words such as ``hot,'' ``stocks,'' ``markets,'' as well as all sorts of currency name and so on, in order to focus on the underlying news information
that is supposed to influence that stock.  This procedure corresponds to filtering out the endogenous component underlying the information flow and price generating process.  Thus, for ``Toyota,'' for example, out of the original $100$ topics, we are left with $26$ useful topics to work with that are associated with the term.

The relative importance of each topic in explaining trading volume of a given stock
is determined by a simple LASSO regression \cite{24,25,26} with positive constraints:
\begin{equation}
Vol(t) = \sum^{K}_{k = 1}w_k I_k(t) + \epsilon(t)~,
\label{lassoregr}
\end{equation} 
where $Vol(t)$ is the normalized trading volume at time $t$.  
Normalization of the trading volume is performed by dividing volume
by the median trading volume within a 2 year moving window 
(boundary values are set to the nearest non-zero value).  The regularized linear regression with mean-squared error provides a robust estimation of the relationship between news topics and trading volume in the presence of large bursts
of trading activity and news, so that a larger span of activity sizes can contribute to the determination 
of the regression weights $\{w_k\}$. The regularization parameter used in the LASSO regression
was chosen equal to the mean value of the regularization parameter over one hundred ten-fold cross validations.  Ten-fold cross validation was performed by randomly dividing the entire data set into 
ten subsets and measuring the average mean-squared error of each testing set from the ten-fold cross validation.  This procedure was performed multiple times  to ensure stability of the estimated regularization parameter.  

Because researchers are generally interested in explaining large (or ``abnormal'') market activity, 
we focus our attention on ``peak days,'' defined in terms of the 95th percentile of daily trading volume, so that on 95\% of the days the trading volume was smaller than during the
peak days.  In order to pay equal attention to large market activity across the whole study period (January 2003 to June 2011), we divided the period overall into 17 six-month time windows and identified the ``peak days'' for each of the 17 time windows separately.  The sequence of peak days is shown in Fig. 4. For each term such as ``Toyota,'' the fraction of the corresponding estimated news volume that can be explained by each topic via regression (\ref{lassoregr}), restricting our attention to only the news volume found on ``peak days,'' is referred to as the ``fraction of volume explained'' (FVE).  In this article, we only use topics that obtained FVE values 
larger than 0.5\%.  For example, this method determines $K=9$ out of $26$ topics as being
useful for ``Toyota.''  Table 1 provides a list of these 9 topics and their individual FVEs for ``Toyota.''  Inspections of this list shows that our procedure yields sensible results, and unimportant topics such as ``Formula One'' shown in Fig. 3 are correctly pruned out.

FIG. 5 compares the observed trading volume with the fitted trading volume using regression (\ref{lassoregr})
(without the residual term $\epsilon(t)$) for four stocks: Toyota, Yahoo, Best Buy, and BP.   
While some parts exhibit a good match, other parts show some discrepancy.
To quantify the quality of the regression and explanatory power of the topic decomposition, 
we focus on the ``peak days'' previously defined and shown in Fig. 4. We define a success
if the predicted volume is at least equal to 10\% of the observed trading volume for a given peak day subtracting the constant value estimated via regression. The fraction of peak days among the total number peak days over the entire period from January 2003 to June 2011 whose volume is successfully accounted for in this sense is referred to as the ``fraction of peaks explained'' (FPE). We obtain the following values: FPE=0.27 (the total number of explained peak days is 32 out of 119) for Toyota, FPE=0.70 (the total number of explained peak days is 83) for Yahoo, FPE=0.51 (the total number of explained peak days is 61) for Best Buy, and FPE=0.43 (the total number of explained peak days is 51) for BP.  

The quality of our regression exercise can be further assessed by comparing the results with those obtained using reference nulls. Specifically, we swap the news associated with different companies. For example, we use
the news records associated with BP and use the extracted topics  
in regression (\ref{lassoregr}) in order to explain the trading volume of Yahoo (left panel of Fig. 6) and use the news record associated with Yahoo to explain the trading volume of Best Buy (right panel of Fig. 6).  This corresponds to modifying only step (1) in the flowchart shown in Fig. 2, while
all the other steps remain the same. As seen in Fig. 6 the explanatory power decreases
considerably, as for instance illustrated by the fact that the FPE is exactly $0$ in both cases. This substantial decrease in explanatory power is found in all our tests and confirms that our regressions done
at the daily scale perform well in pruning out unimportant topics and identify the relevant ones. 
Obviously, (i) if the two companies for which news records are swapped have some commonalities (e.g., they are engaged in merger talks), or (ii) if they always disclose their earnings reports on exactly the same date throughout the entire observation period, then some topics found for one stock would explain the trading activity of the other, but this is rarely the case.

\subsection*{Results}

We applied the methodology introduced in the previous section to the 206 companies listed in the S\&P 500 US stock index for which there were more than 5,000 news records during the period from January 2003 to June 2011.  Fig. 7 plots the FPE metric as a function of the number of news records for these 206 stocks.  

Over the set of the 206 analyzed US stocks, 715 topics were found to have
a significant impact on trading activity. Recalling that the logic of topic models, as highlighted by \cite{20}, is that corpus meanings are organized in topics that share global multinomial word distributions, a convenient way to visualize the similarities between topics is to use network graphs. 
We therefore construct networks with topics as nodes, and  a link between two topics exists when the 
Jensen-Shannon Divergence (JSD) \cite{JSD}
between the two corresponding topic distributions is smaller than $0.5$.
The size of a node is set to be proportional to the ``fraction of volume explained'' (FVE)
by that topic and the thickness of a link is equal to $1$ minus the JSD metric for the two linked topics.
Each topic is labeled by its top three most frequent words,
as quantified by the topic distribution, together with the company's name.  
We also depict all the companies name with a fixed size of 0.5 and connected all their selected topics with them where the edge strength was set to their FVE value.  
The networks are depicted using the Force Atlas algorithm using the freely available software Gephi(https://gephi.org/)

Fig. 8 shows the network of topics for the two stocks Microsoft and Yahoo.  Both have topics 
reflecting earning reports and exhibit features that reflect a potential merger deal.  From the
node sizes (proportional to their FVEs),  one can clearly see that the potential merger deal between the two 
companies had more impact on Yahoo's stock than on Microsoft's stock.
This is in agreement with the fact that Yahoo was facing difficulties in 2009. 
This demonstrates another useful property of our method, which is that it allows us to quantify and compare the impact of two or more external influences.

Fig. 9 shows the whole network of all the topics extracted by our method for the 206 stocks we focus on.  The network can be viewed as consisting of the ``mainland'' and more isolated ``islands.'' 
The mainland is made up of all the connections between topics produced by words reflecting earnings reports (``profit,'' ``earning,'' ``share,'' ``pct (short for percent)''), credit ratings (``rating,'' ``debt,'' ``credit''), merger deal (``merger,'' ``deal''), and the financial crisis (``crisis,'' ``financial''). In order to better discern some of the major ``islands,'' 
Fig. 10 presents six zooms on the domains indicated by the arrows in Fig. 9.
The observed clusters of company names and words representing the topic distributions
confirm that our method successfully extracted the correct information.  
Note that all the word contents of the constructed topic distributions have financial and/or economic meaning that carry useful information from the point of view of an investor and can be surmised
to indeed have an impact on the future earning of the firms. We refer in particular
to the following word contents:  ``earning reports,'' ``retailers profits,'' ``drug patents,'' ``national defense budget,'' 
``new products,'' ``merger deal,'' ``global recession,'' ``natural disasters,'' and so on. 

To assess further the quality of our regressions, we manually ``read'' all the 715 topic distributions,
identifying the underlying news records that contained each topic to some extent.  As could be suspected 
given our approach, not all topics qualified as conveying meaningful information:
among the 715 topic distributions, we determined that 78 were misspecified.  Those topics were either (1) reflecting news words that were not correctly pruned out by our procedure (such as ``reuters,'' ``users,'' ``click''), (2) market words that were not correctly pruned out (``imbalance'' ,``nyse,'' ``trademark''),
or  (3) incorrect information extracted due to the peculiarity of our data that one news record sometimes contains more than one piece of information news (for instance, this is due to news records that list the top news of the day).  To determine the impact of excluding these miss-extracted topics, blue circles in Fig. 7 show the FPE value excluding these misspecified topics.  We see that the overall FPE value does not change much, supporting
our trust in the robustness of our approach.

The other 89\% (i.e. 637) topic distribution contained relevant information.  To classify these remaining topics, we first combined duplicated topics for each stocks (for instance, the third and ninth topic in table 1 both reflect earnings reports). This leads to 44 broad categories, which are listed in Table 2.  Our method tends 
to put more emphasis on regular reporting about the future earning of the firms, but also successfully extracts peculiar incidents that are suspected to change the course of the future earnings of the firms.  
Summing up all these investigations, we conclude that we have successfully extracted the important pieces of information that influence financial markets.

\section*{Discussion}

In this study, we performed an analysis of more than 24 million news records provided by Thompson Reuters 
and of their relationship with trading activity of the stock of 206 major firms included in the S\&P 500 index.
We showed that the whole landscape of the news that affect stock price movements can be automatically summarized by conducting a simple regularized regression between trading activity and news information pieces decomposed into their ``thematic'' features, with the help of simple topic modeling techniques. Using these methods, not only were we able to extract the pieces of information that synchronize well with trading activity but, as a bonus of the simultaneous regressions, we were also able to estimate and quantify their impact, which is difficult to do otherwise. We also introduced novel ways to visualize the whole landscape of news information 
associated with a basket of stocks by utilizing network visualization techniques.
The examination of the words that are representative of the topic distributions and careful reading of the news records which included that topic to some extent confirmed that our method successfully extracted the significant pieces of information influencing the stock market.   

Our finding of a high explanatory power of news to 
account for stock market trading activity provides insights on the question
raised in the introduction on the nature of the news that may influence stock markets
and how they are digested in stock prices.  
In particular, our results show that large volumes of trading can often be explained
by the flow of  news. In this sense, our results might suggest that ``excess trading"
is not always prevalent, especially when the news are genuinely novel and provide relevant financial information.

One of the reasons for the success of our simple methodology, which does not require
taking into account lag effects or more sophisticated nonlinear dynamics, is probably the high quality of the news sources, which resulted in a high signal-over-noise 
ratio. Specifically, the news that we used are gathered
for professional investors, who incentivize the collecting firm by
paying significant subscription fees. Our study confirms the 
exceptional relevance of such professional financial sources compared with 
other standard textual information such as tweets or blogs. The size of our database
in terms of the number of news records compared with that available from standard newspapers was
also essential for the extraction of the important topics that influence the trading
activity of financial markets. In conclusion, we believe
that our results summarize the major sources of external influences on financial markets
stemming from news information associated with them. Another challenge beyond
explaining trading activity is to explain pricing and financial valuations in general, 
using the extended universe of news, topics, and their networks. This is left for future work.

\section*{Acknowledgments}
The authors are grateful to Vladimir Filimonov, Georges Harras, and Ryan Woodard for helpful discussions and comments concerning this work.  Ryohei Hisano is partially supported by funding from the Japanese Student Services Organization through a scholarship titled ``Scholarship for Long-term Foreign Studies-2010.'' Tsutomu Watanabe is supported by funding from JSPS Grant-in-Aid for Scientific Research (24223003).

\bibliography{HisanoBib}

\section*{Figure Legends}

\newpage
\begin{figure}
\begin{center}
\includegraphics[width=5in,bb=  98 448 705 801]{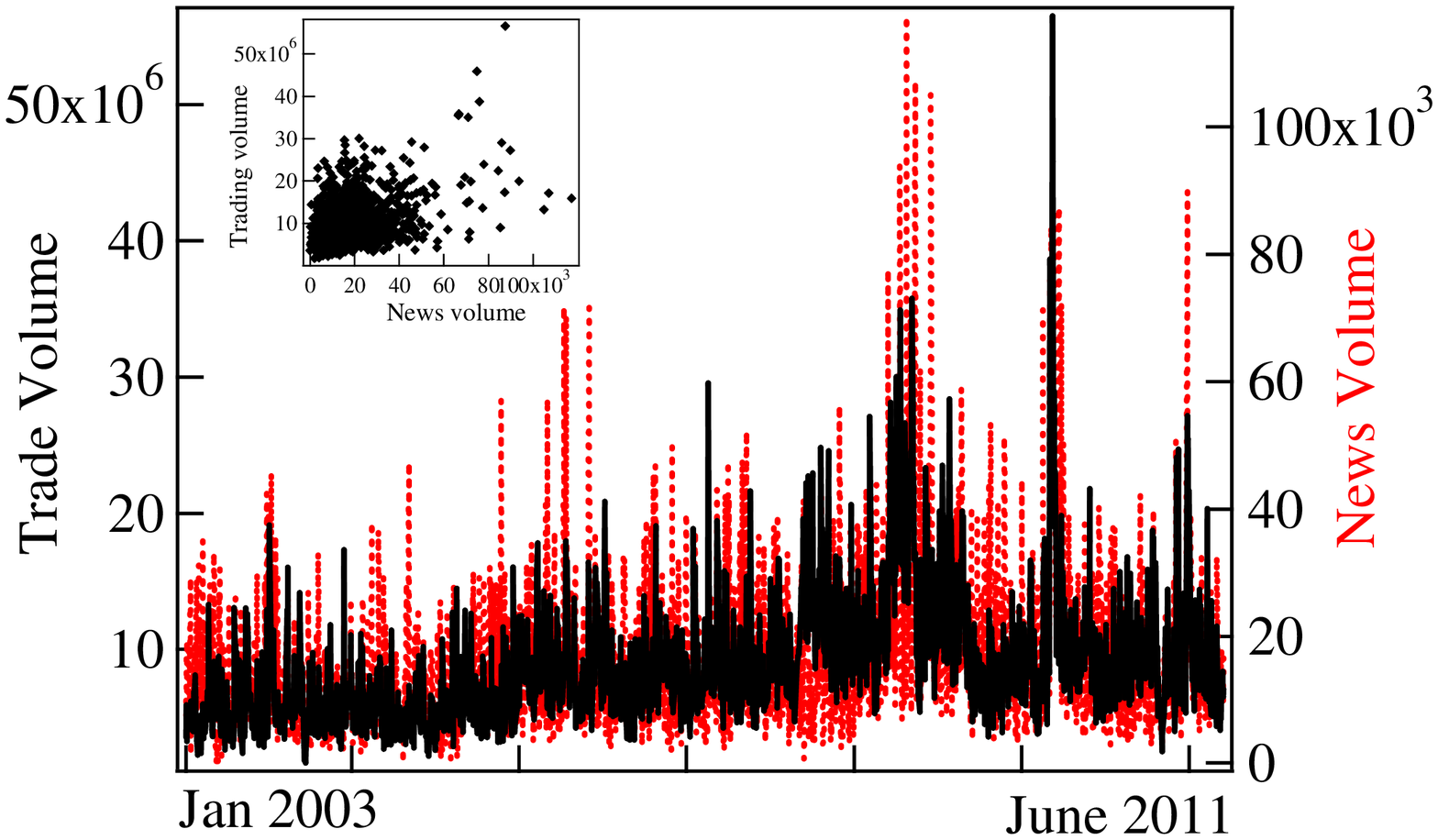}
\end{center}
\caption{{\bf Comparison between the time evolution of trading volume and 
aggregate news volume for Toyota.}  Black continuous line plots trading volume and red dashed line plots aggregate news volume. The inset plots
the trading volume as a function of the concomitant news volume.}

\end{figure}

\newpage
\begin{figure}
\begin{center}
\includegraphics[width=1\hsize,bb=  33 7 682 539]{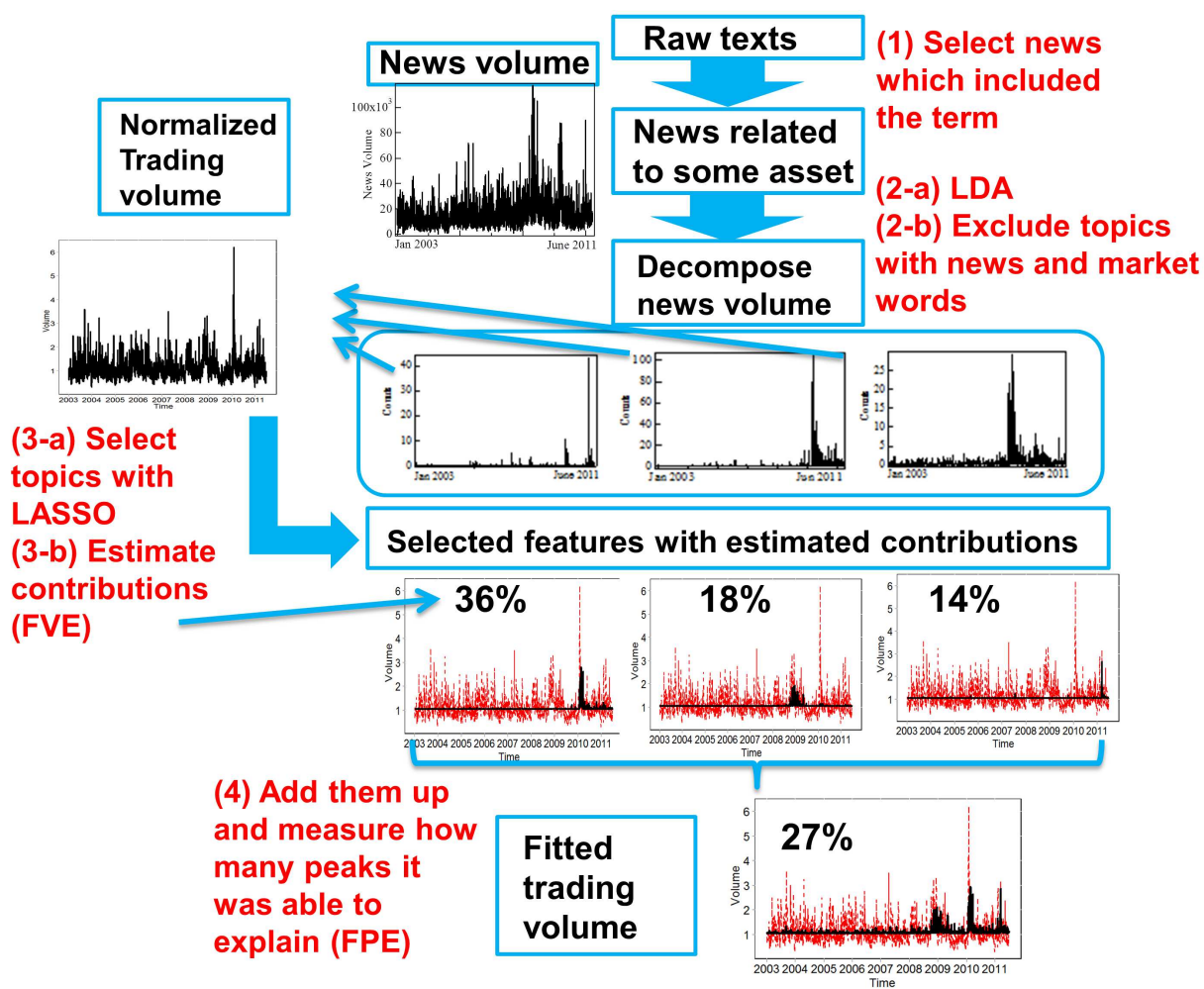}

\end{center}
\caption{{\bf Flowchart summarizing the procedure followed in our analyses.}  The number in parentheses indexes the step.  Step (1) selects the news records associated with a given term, here the name of a company, such as Toyota. Step (2-a) applies the Latent Dirichlet Allocation (LDA) that decomposes any document as a mixture of different topics. Step (3-a) implements a constrained LASSO regression. The percentage shown in step (3-b) denotes the estimated impact of each topic. The percentage shown in step (4) is the ``fraction of (trading volume) peaks explained'' (FPE) by news, which is our metric to assess the quality of our methodology (see text).} 
\end{figure}

\newpage
\begin{figure}
\begin{center}
\includegraphics[width=1\hsize,bb= -0 -0 529 524]{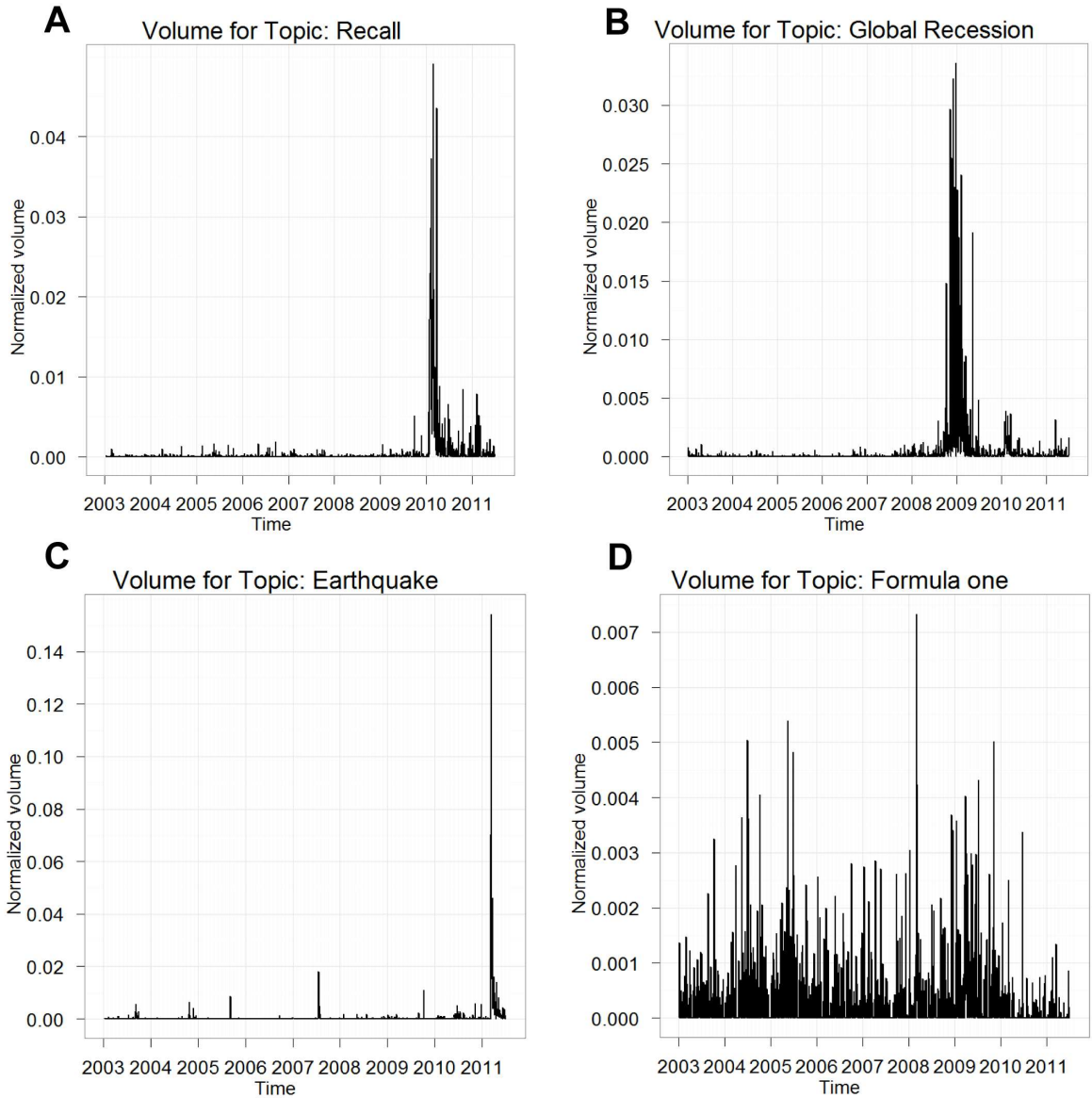}

\end{center}
\caption{{\bf Selected topic learned by LDA for Toyota.} Selected topics learned by LDA and the associated news volume estimated using equation (1) for the term ``Toyota.'' The top three words for these topics were: (A) Toyota, recall, safety; (B) financial, crisis, economy; (C) Japan, production, earthquake; (D) team, F1, race.}
\end{figure}

\newpage
\begin{figure}
\begin{minipage}{\hsize}
\begin{center}
\includegraphics[width=1\hsize,bb=0 0 648 648]{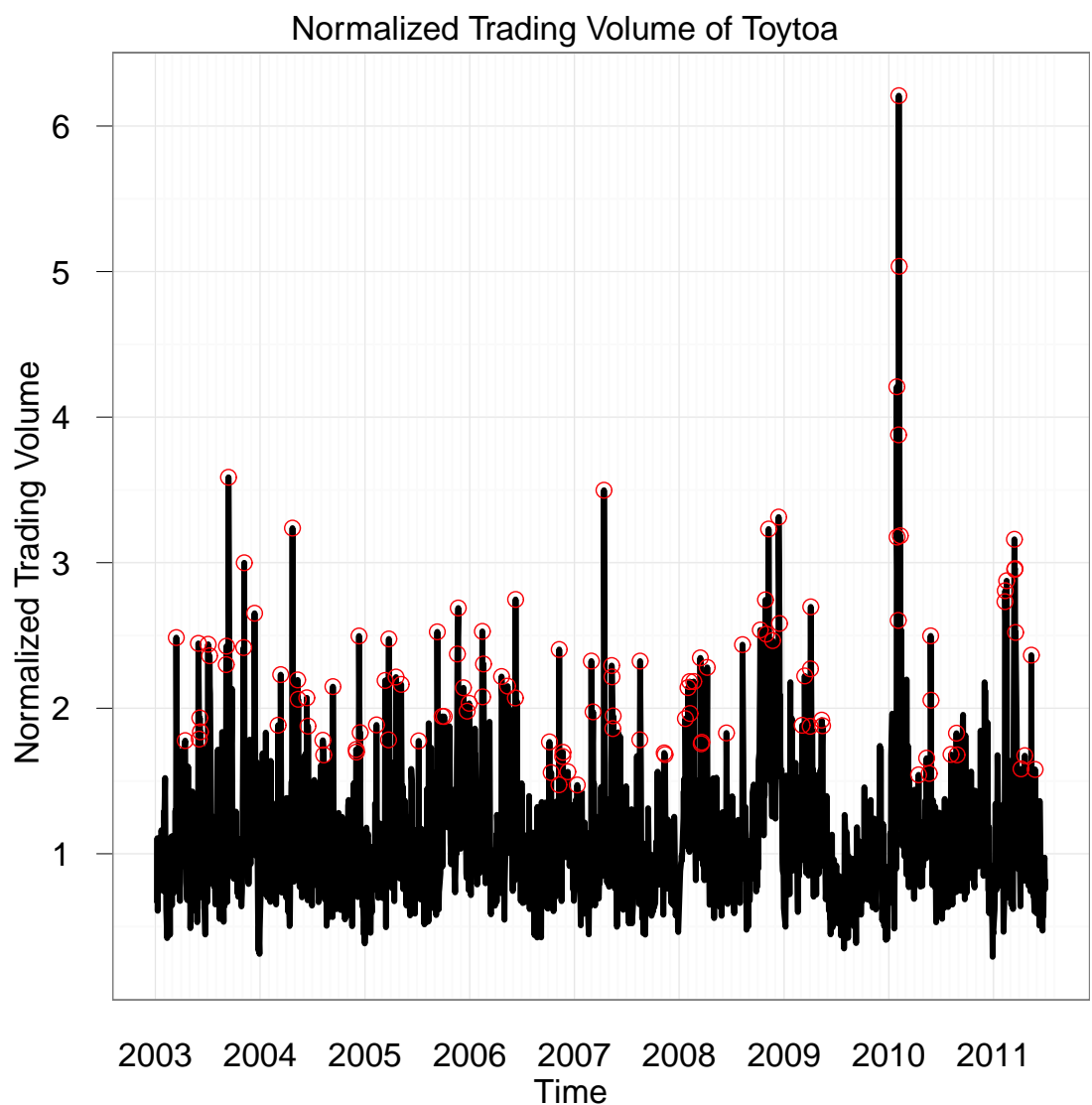}
\end{center}
\caption{{\bf Pictorial illustration of ``peak days'' of normalized trading volume.} The black line shows the de-trended trading volume of Toyota stock for the period from January 2003 to June 2011. The red dots indicate the ``peak days'' selected by the method described in the text. There are 119 ``peak days'' for the entire period from January 2003 to June 2011.}
\end{minipage}
\end{figure}

\newpage
\begin{figure}
\begin{center}
\includegraphics[width=0.8\hsize,bb=  67 16 569 526]{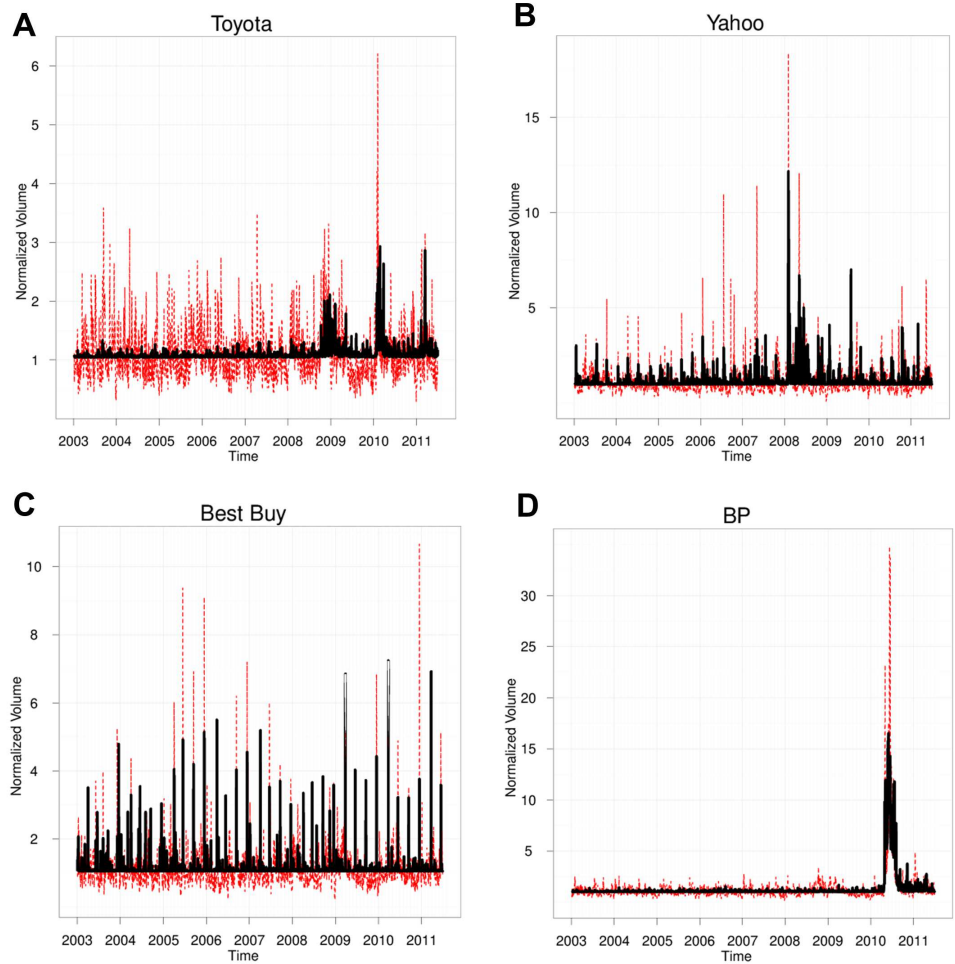}

\end{center}
\caption{{\bf Comparison between estimated and actual trading volume.} Estimated (red dashed line) and actual (black continuous line) trading volume for the four companies: (A) Toyota, (B) Yahoo, (C) Best Buy, and (D) BP.  The number $K$ of sufficient selected topics is 9 for Toyota, 4 for Yahoo, 3 for Best Buy, and 5 for BP.}
\end{figure}

\newpage
\begin{figure}
\begin{center}
\includegraphics[width=0.8\hsize,bb=  67 283 573 526]{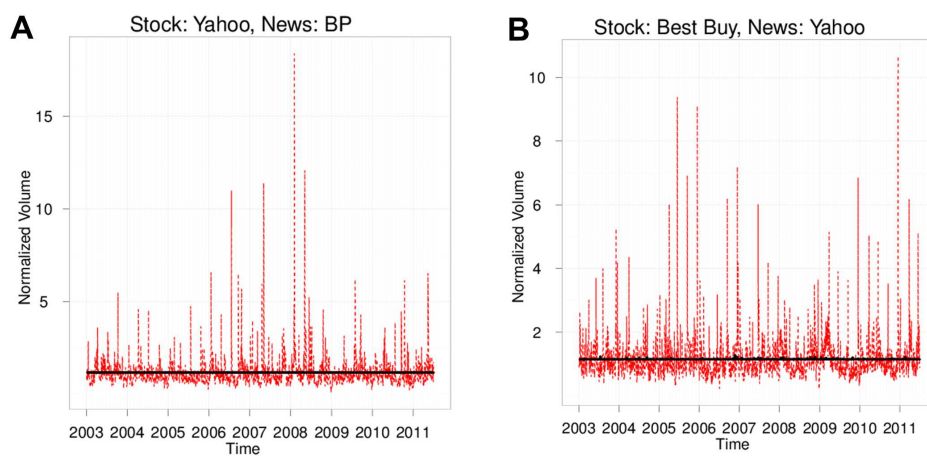}
\end{center}
\caption{{\bf Result of stress testing.} (A) Comparison between the estimated and actual trading volume when using topics from BP when trying to explain Yahoo trading volume.  (B) Comparison when 
using topics from Yahoo when trying to explain Best Buy trading volume.  Notice the much reduced quality of the regressions compared with those presented in Fig. 6,
illustrated by their FPEs, which are exactly $0$ in both cases.}
\end{figure}

\newpage
\begin{figure}
\begin{minipage}{\hsize}
\begin{center}
\includegraphics[width=1\hsize,bb=16 9 412 387]{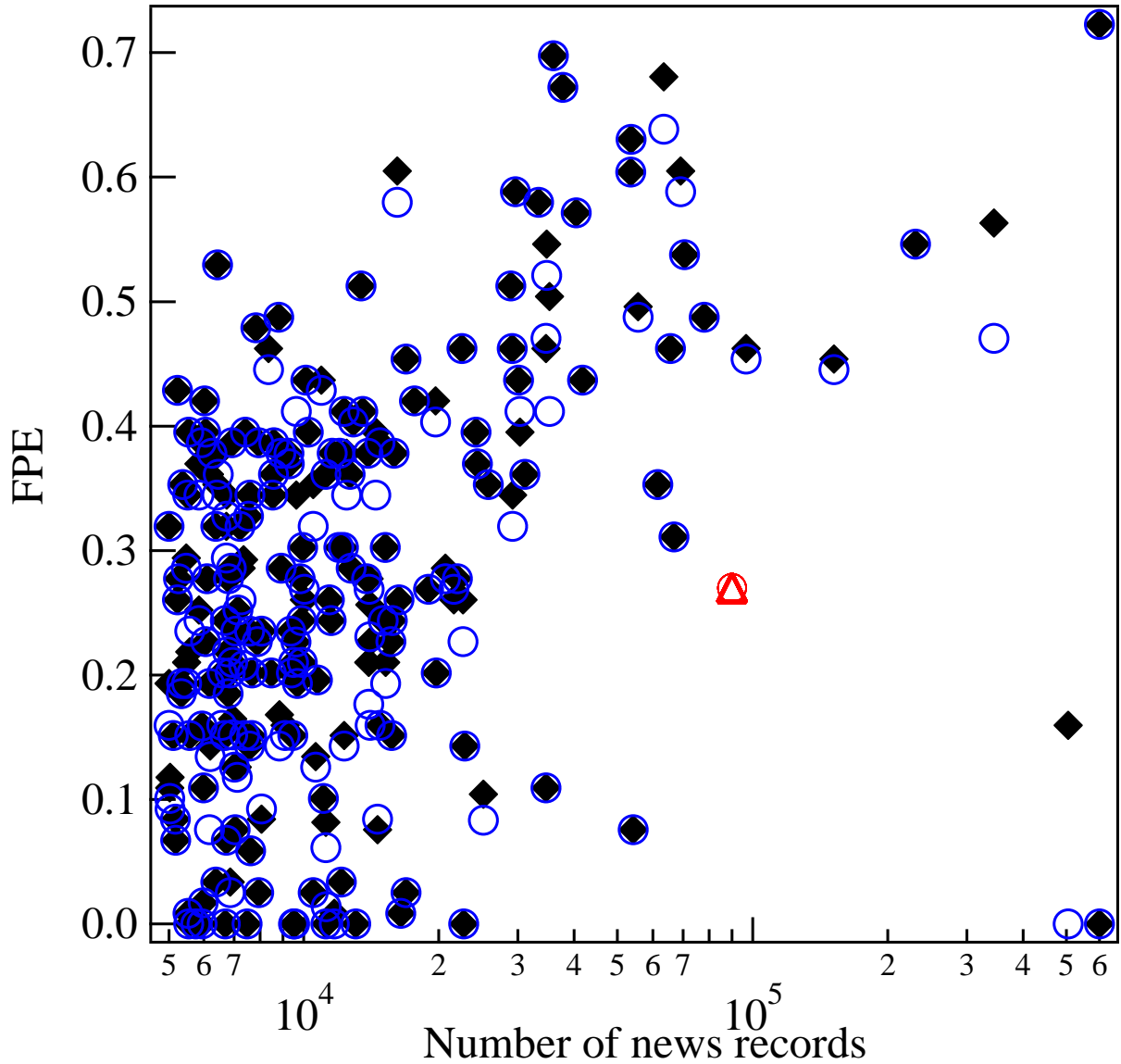}
\end{center}
\caption{{\bf Relationship between FPE and number of news records.} The ``fraction of peaks explained'' (FPE) as a function of the number of news records for the 206 stocks in the S\&P 500 
for which there were more than 5,000 news records
during the period from January 2003 to June 2011. Black diamond shows the FPE value using the 715 topics extracted from our procedure.  Blue circle shows the FPE value restricting the number of topic distributions to 637 after manual reading.  The data point for Toyota, which as a foreign company of course is not a component of the S\&P 500, has been added and is shown as the red triangle and circle.}
\end{minipage}
\end{figure}

\newpage
\begin{figure}
\begin{minipage}{\hsize}
\begin{center}
\includegraphics[width=1\hsize,bb= -0 -0 779 539]{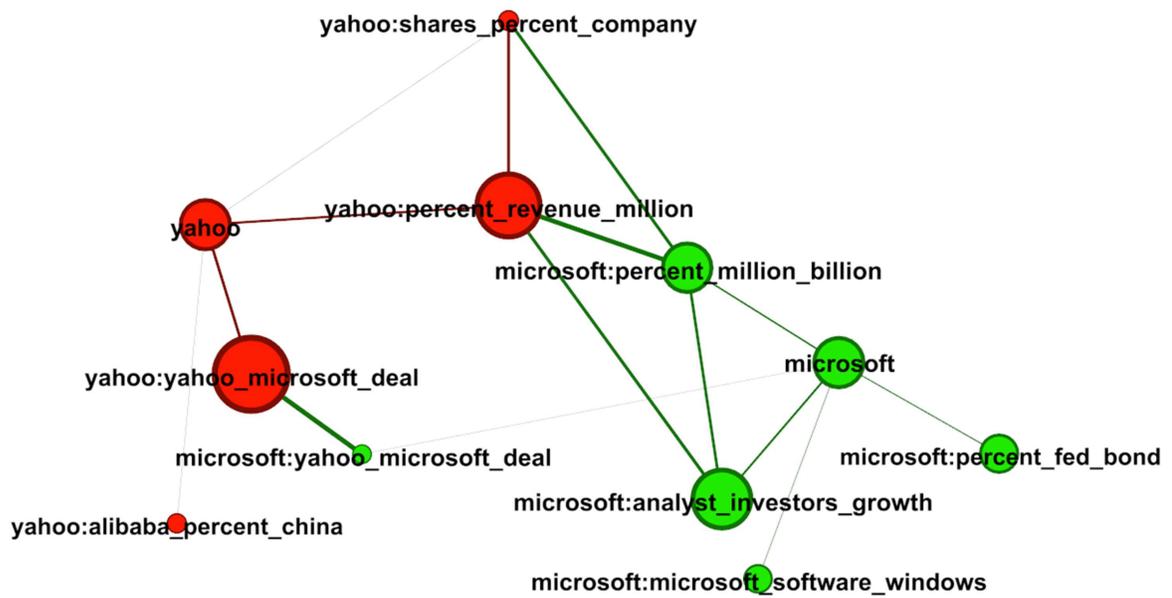}
\end{center}

\end{minipage}
\caption{{\bf Network extracted for Microsoft and Yahoo.}  Nodes are topics and links
between two topics quantify the degree of similarity associated with their word distributions.}
\end{figure}

\newpage
\begin{figure}
\begin{center}
\includegraphics[width=1\hsize,bb=  4 13 692 534]{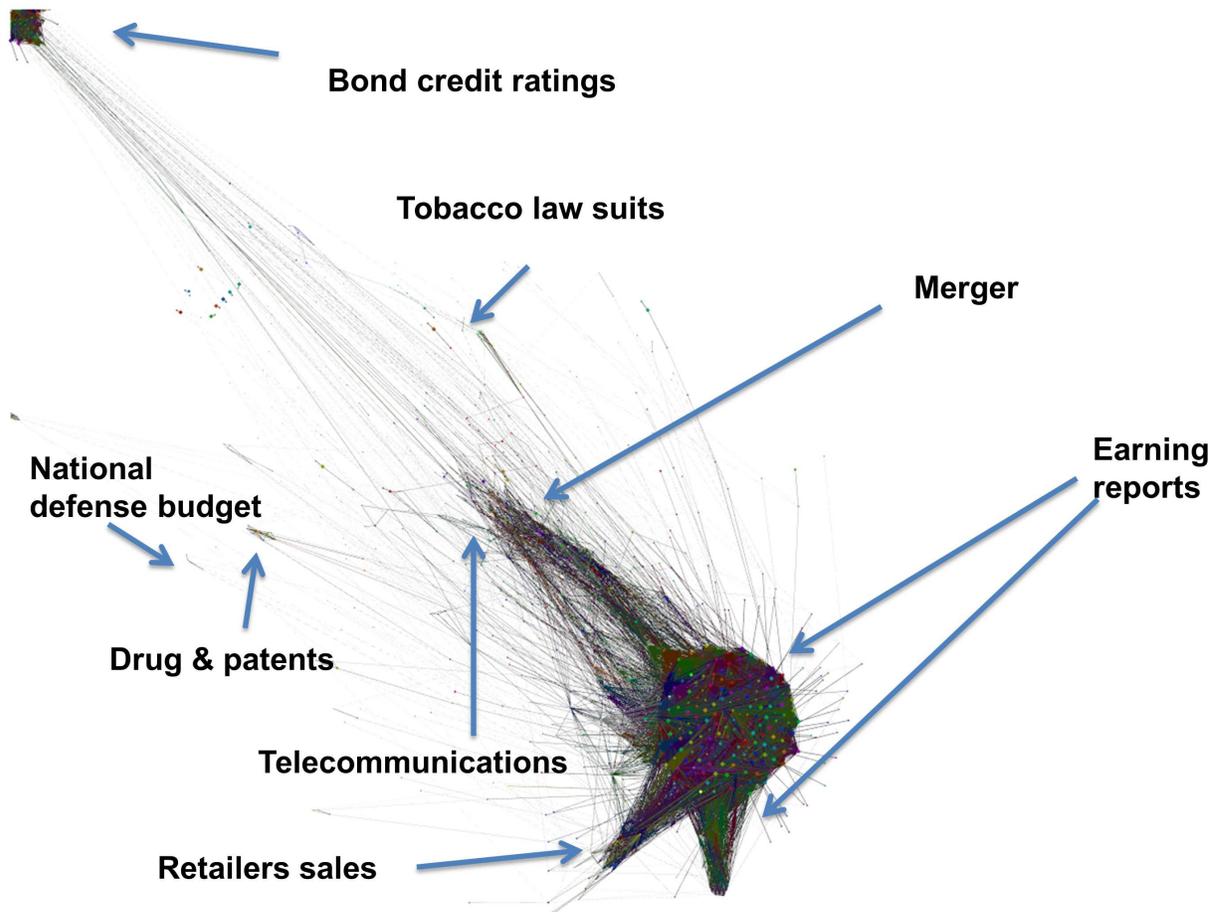}
\end{center}

\caption{{\bf Network of topics extracted for the 206 US companies.} The links between two topics quantifying the degree of similarity associated with their word distributions,
as explained in the text. The six red arrows depict the zones that are magnified in Fig. 10.}
\end{figure}

\newpage
\newpage
\begin{figure}
\begin{center}
\includegraphics[width=0.7\hsize,bb= 0 19 540 780]{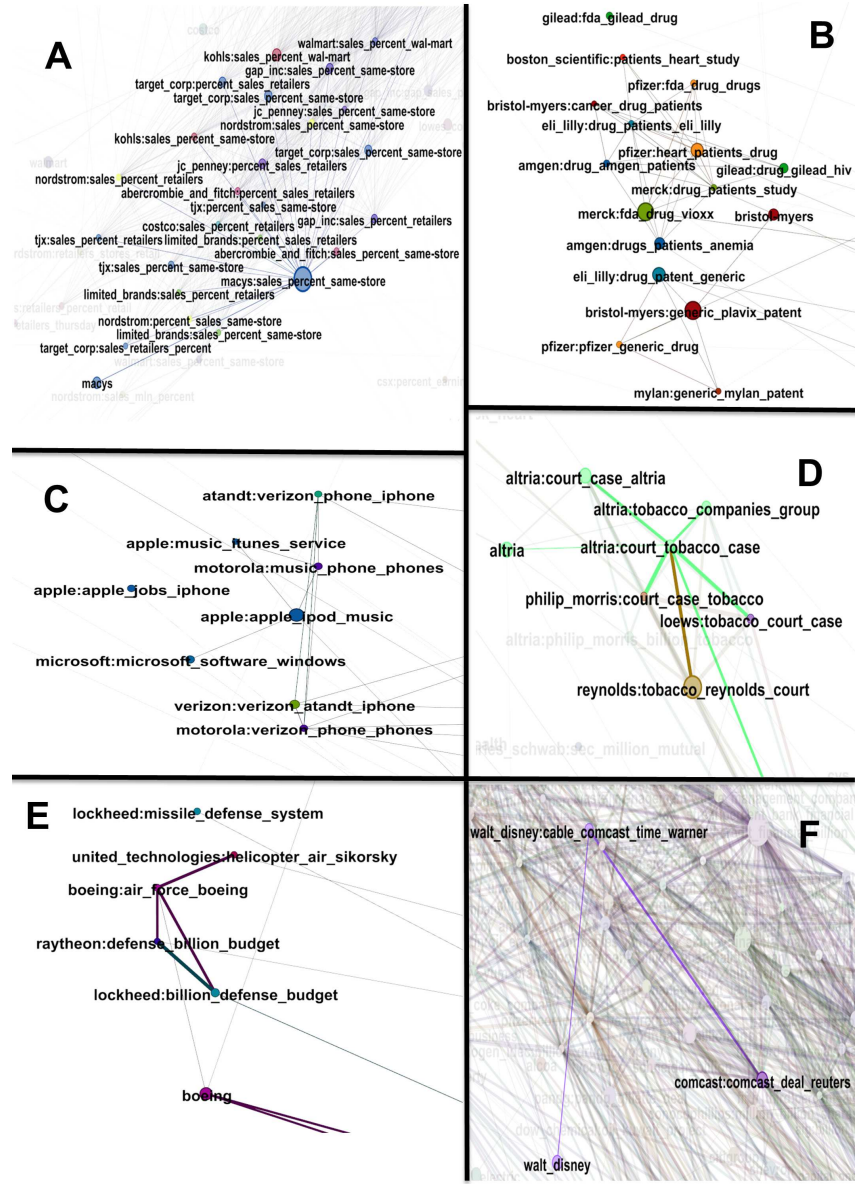}
\end{center}
\caption{{\bf Magnifications of Figure9.} Six magnifications of the ``islands'' indicated by the arrows in the network of topics shown in Fig. 10,
with links between two topics quantifying the degree of similarity associated with their word distributions.
Each node is accompanied by the name of the company and its top three most frequent words,
as quantified by the topic distribution. 
The size of a node is set to be proportional to the ``fraction of volume explained'' (FVE)
by that topic and the thickness of a link is equal to $1$ minus the JSD metric for the two linked topics. 
Panel (a) shows the network associated with retail sales of clothing companies; panel (b) that associated with 
drug and patents; panel (c) that associated with products in telecommunication business;
panel (d) that associated with tobacco law suit; panel (e) that associated with national defense budget;
panel (f) that associated with the potential Comcast Disney merger in 2004.}
\end{figure}

\newpage
\section*{Tables}

\begin{table}[!ht]
\caption{
\bf{List of the 9 selected topics for ``Toyota''.} Their estimated ``fraction of volume explained'' (FVE) are shown as well.  
Topic distributions are summarized with their top most frequent 3-5 words.  For a full description of the topic distributions, see the supporting information. 
}
\begin{tabular}{|c|c|c|}
\hline
Rank & Top words & FVE  \\

\hline
1 & (toyota,recall,safety) & 0.36\\
\hline
2 & (financial,crisis,economy) & 0.18\\
\hline
3 & (profit,yen,billion) & 0.15\\
\hline
4 & (japan,production,earthquake) & 0.15\\
\hline
5 & (steel,percent,nippon,prices,demand) & 0.05\\
\hline
6 & (economy,japan,percent) & 0.05\\
\hline
7 & (hybrid,cars,car) & 0.02\\
\hline
8 & (car,fiat,euro) & 0.01\\
\hline
9 & (pct,company,earnings) & 0.01\\

\hline
\end{tabular}
\begin{flushleft}
\end{flushleft}
\label{tab:label}
 \end{table}

\begin{table}[!ht]
\caption{
\bf{Table summarizing the content of our extracted topics.} 
The classification was created by manual reading of the underlying news records that included the topic, at least to some extent.  The right column shows the number of stocks that contained that topic out of the 206 analyzed stocks.  See text for more information.}
\begin{tabular}{|c|c|c|}
\hline
Number & Classification & Number of stocks  \\

\hline
1 & Quarterly earnings & 187 \\
\hline
2 & Bond credit ratings & 57 \\
\hline
3 & Merger and acqusition &	41 \\
\hline
4 & Sales (revenue) of products (stores) &	25 \\
\hline
5 & Lawsuit &	17 \\
\hline
6 & Financial crisis &	17 \\
\hline
7 & Top management(board / gossip)	& 16 \\
\hline
8 & Product information	& 15 \\
\hline
9 & Drugs(patents / controversy / approval)	& 12 \\
\hline
10 & Business deal & 10 \\
\hline
11 & Corporate bond &	8 \\
\hline
12 & Flawed accounting / Insider trading / Late trading / SEC &	8 \\
\hline
13 & Recall &	6 \\
\hline
14 & Bankruptcy (of other firms) &	6 \\
\hline
15 & Shortlist/ Takeover / Selling own stocks &	5 \\
\hline
16 & Energy prices &	5 \\
\hline
17 & Legislation / Regulation / Bill &	4 \\
\hline
18 & Natural disaster &	4 \\
\hline
19 & BP oil spill &	4 \\
\hline
20 & National defense &	3 \\
\hline
21 & Strike &	3 \\
\hline 
22 & Outbreak &	3 \\
\hline
23 & Medical industry &	3 \\
\hline
24 & Central America economy &	3 \\
\hline
25 & Dividend &	2 \\
\hline
26 & Licensing (airwaves / licensing in middle east) &	2 \\
\hline
27 & Lay off &	2 \\
\hline
28 & Power plant &	2 \\
\hline
29 & Oil refinery &	2 \\
\hline
30 & Building pipeline &	2 \\
\hline
31 & Precious metal &	2 \\
\hline
32 & Media industry &	2 \\
\hline
33 & Fast food industry &	2 \\
\hline
34 & India economy &	2 \\
\hline
35 & IPO (of related firms) &	1 \\
\hline
36 & Business plan & 1 \\
\hline
37 & Blackout &	1 \\
\hline
38 & Subplime loan problem &	1 \\
\hline
39 & Government bailout &	1 \\
\hline
40 & Gas &	1 \\
\hline
41 & Environmental issue &	1 \\
\hline
42 & Steel &	1 \\
\hline
43 & Online education business &	1 \\
\hline
44 & Middle east economy &	1 \\
\hline
\end{tabular}
\begin{flushleft}
\end{flushleft}
\label{tab:label}
\end{table}
\end{document}